\documentclass{article} 
\usepackage{iclr2016_conference,times}

\usepackage{graphicx}
\usepackage{xspace}
\usepackage{rotating}
\usepackage{multirow} 
\usepackage{cite}
\usepackage{url}
\usepackage{verbatim}

\usepackage{amsmath}
\usepackage{amssymb}
\usepackage{bm} 

\usepackage[marginclue,inline,final]{fixme}

\newcommand{\lspan}[1]{\ensuremath{\mathrm{span}}}

\newcommand{\reffig}[1]{Fig.~\ref{#1}}

\newcommand{\refeq}[1]{Eq.~\ref{#1}}

\makeatletter 
\DeclareRobustCommand\onedot{\futurelet\@let@token\@onedot}
\def\@onedot{\ifx\@let@token.\else.\null\fi\xspace}

\makeatother

\usepackage{hyperref}
\usepackage{url}
\usepackage{floatrow}
\newfloatcommand{capbtabbox}{table}[][\FBwidth]

\newcommand{\mycomment}[1]{}

\renewcommand{\cite}[1]{\citep{#1}}
\newcommand{\reftable}[1]{Table.~\ref{#1}}

\newcommand{\ba}{\begin{eqnarray}}
\newcommand{\ea}{\end{eqnarray}}

\begin{document}
	\title{Pushing the Boundaries of Boundary Detection using Deep Learning}
	\author{Iasonas Kokkinos\\
	Center for Visual Computing\\
	CentraleSup\'elec and INRIA \\
	Chatenay-Malabry, 92095, France\\
	\texttt{\{iasonas.kokkinos\}@ecp.fr} \\
	}
	\maketitle
		
\begin{abstract}
	In this work we show that adapting Deep Convolutional Neural Network training to the task of boundary detection can result in substantial improvements over the current state-of-the-art in boundary detection. 
	
Our contributions consist firstly in combining a careful design of the loss for boundary detection training, a multi-resolution architecture and training with external data to improve the detection accuracy of the current state of the art.
When measured on the standard Berkeley Segmentation Dataset, we improve theoptimal dataset scale F-measure from 0.780 to 0.808  - while human  performance is at 0.803. We further improve performance to 0.813  by combining deep learning with grouping, integrating the  Normalized Cuts technique within a deep network.

We also examine the potential of our boundary detector in conjunction with the  task of semantic segmentation and demonstrate clear improvements over state-of-the-art systems. 
Our detector is fully integrated in the popular Caffe framework and processes a 320x420 image in less than a second.


\end{abstract}

\section{Introduction}

Over the past three years Deep Convolutional Neural Networks (DCNNs)
\citet{LeCun1998}
 have  delivered compelling results in high-level vision tasks,  such as  image classification \citep{KrizhevskyNIPS2013, SEZM+14, simonyan2014very, szegedy2014going, papandreou2014untangling} or  object detection \citep{girshick2014rcnn}.
Recent works have also shown that DCNNs can equally well apply to pixel-level labelling tasks, including semantic segmentation 
\citep{LongSD14,Chen2015iclr} or  normal estimation \citep{NIPS2014_5539}. A convenient component of such works is that the inherently convolutional nature of DCNNs allows for simple and efficient `fully convolutional' implementations \citep{SEZM+14,NIPS2014_5539,Oquab13,LongSD14,Chen2015iclr}.


Our focus on this work is the low-level task of boundary detection, which is one of the cornerstone problems of computer vision. Segmentation can be  considered to be an ill-posed problem, and multiple solutions can be considered plausible, depending on the task at hand - for instance when playing chess we think of a checker board in terms of 64 regions, but when carrying it we treat it as a single object. This is reflected in the inconsistency of human segmentations, illustrated in \reffig{fig:cover}.

As detailed in \citet{berkeley11} we can `benchmark' humans against each other, by comparing every annotator to the `committee' formed by the rest: if a user provides  details that no committee member has provided these count as false positives, while if a user misses details provided by a committee member, these count as misses. Aggregating this information over different annotators yields the recall and precision of humans, which are in turn summarized in terms of their f-measure, namely their geometric mean. When evaluated on the test set of Berkeley Segmentation Dataset (BSD) humans  have an F-measure of $0.803$, which is indicative of the difficulty of the task. 

This difficulty may be substantially diminished if we consider segmentation as an intermediate to a specific task, such as object detection; it has been shown for instance in \citet{ZhuTMD15} that when asking users to provide a  label to every region the F-measure of human annotators rapidly increases from 0.8 to 0.9. Still, when considering the segmentation problem in its largest generality, namely as a mid-level task serving detection, tracking, counting, or even grasping and touching, the ambiguity of the labelling most naturally increases. 
 
Despite the inherent  difficulty of the problem, progress in boundary detection has been consistently narrowing the gap between human and machine performance, as measured 
 Our system yields a higher F-measure than humans: when using a common threshold for the whole dataset (Optimal Dataset Scale -ODS) our system's F-measure equals $F=0.813$, while when an oracle sets the threshold per image (Optimal Image Scale -OIS) we obtain $F=0.8308$.

 


\newcommand{\imroot}{./}
\newcommand{\scln}{.17}
\newcommand{\sclhl}{.09}
\newcommand{\sclq}{.05}


  	

As in all works following the introduction of human-annotated datasets \cite{KYCZ03,MFM04},  e.g. \cite{DTB06,berkeley11,Ren08,Kokkinos10,Ren12,sfs}, 
we use machine learning to optimize the performance of our boundary detector. Recent works \cite{shi15,aistat,iclrbnd} have shown hat DCNNs yield substantial improvements over flat classifiers; the Holistic Edge Detection approach of \citet{HED} recently achieved dramatic improvements over the previous state-of-the-art, from an F-measure of $0.75$ to $0.78$, while keeping computation efficient, requiring 0.4 seconds on the GPU; additional dataset augmentation yielded an F-measure of $0.79$. 

\mycomment{
In this work we push the performance to show the we can surpass the performance of humans on this task by employing a multi-resolution
DCNN architecture, trained in a manner that accommodates the challenges of the boundary detection task. 
}

In this work we make contributions in three fronts: firstly we improve the deep learning algorithms used for boundary detection, 
secondly we incorporate classical ideas from grouping into the problem and thirdly we exploit our detector to improve the higher-level tasks of semantic segmentation and region proposal generation.
We detail these three advances in the following three sections.

\newcommand{\scl}{.16}
 \begin{figure}
\begin{centering}
 \includegraphics[width=\scl\linewidth]{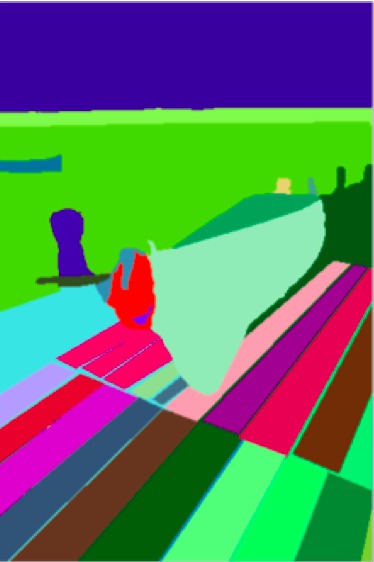}
 \includegraphics[width=\scl\linewidth]{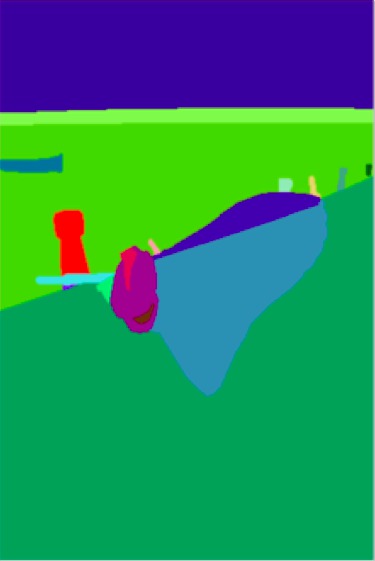} 
 \includegraphics[width=\scl\linewidth]{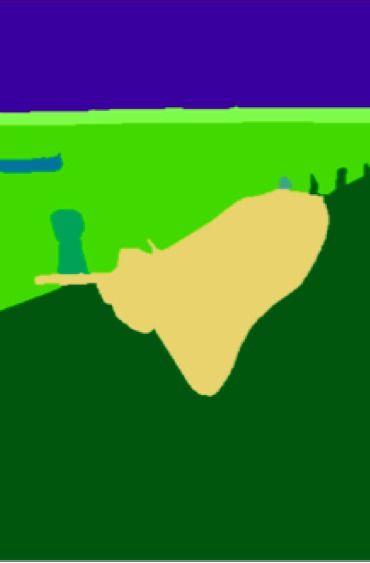} 
 \includegraphics[width=\scl\linewidth]{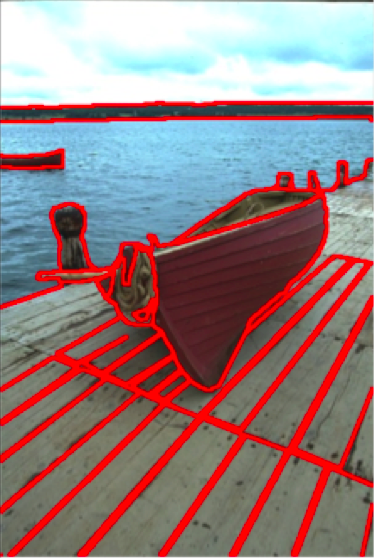}
 \includegraphics[width=\scl\linewidth]{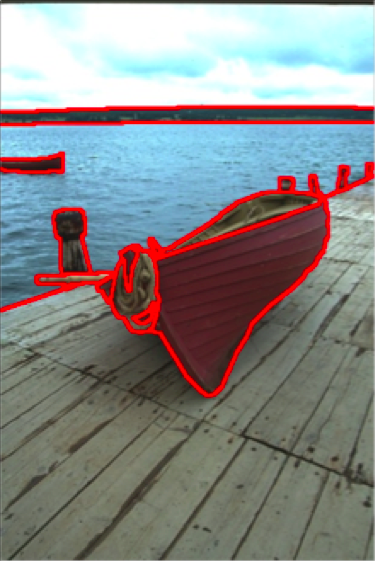} 
 \includegraphics[width=\scl\linewidth]{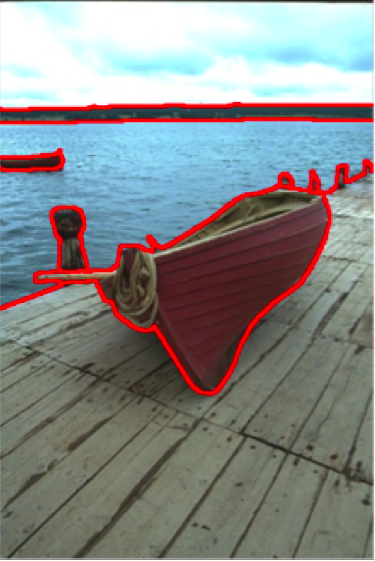} 
 \end{centering}
 \caption{Ground-truth segmentations provided by different annotators for an image from the BSD dataset, and associated boundary maps. The evident lack of agreement among humans is reflected in a low F-measure of human annotators on the task, $F=0.803$. Our system delivers $F=0.813$. 
  	 \label{fig:cover}}
 \end{figure}

\section{HED and DSN training}

\begin{figure*}
 \vspace{-1cm}
 \centerline{	\includegraphics[width=.7\linewidth]{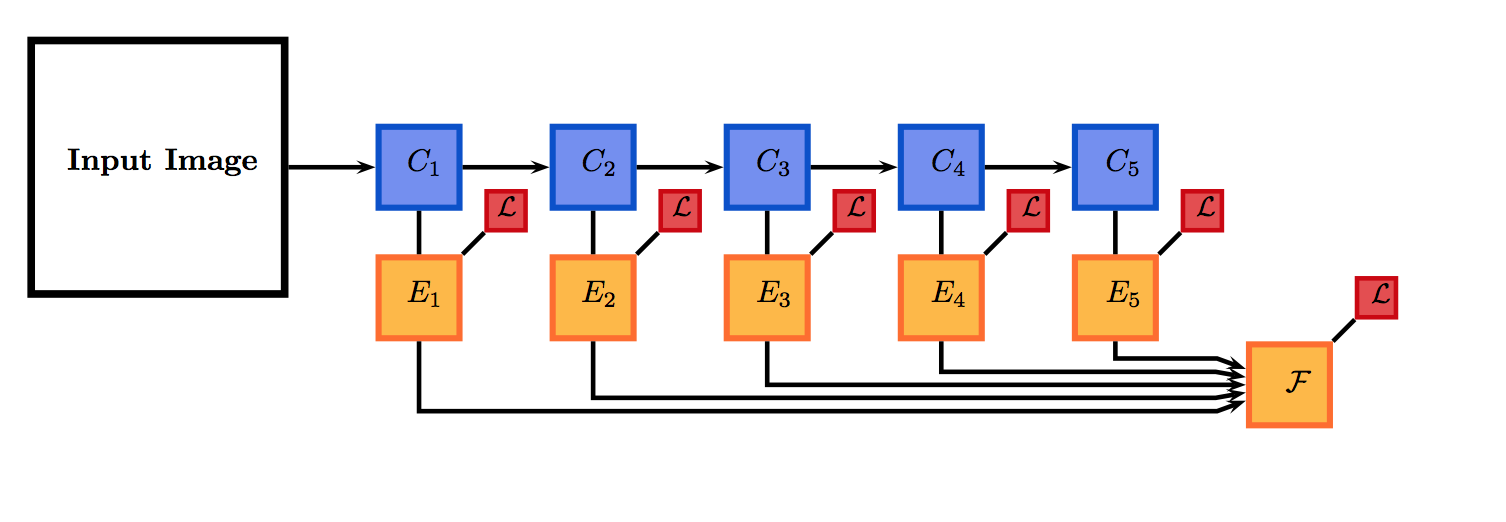}
}
\vspace{-.5cm}
 \caption{HED/DSN training architecture: every intermediate layer of a DCNN (shown blue) is processed by a side layer  (shown in orange) which is penalized by a loss function $\mathcal{L}$. The intermediate results are combined in a late fusion stage, which is again trained with the same loss function.
 	 \label{fig:hedtrain}}
 \end{figure*}

We start from a brief presentation of the `Holistic Edge Detection' (HED) work of \citet{HED} as it serves as a starting point for our work. 
HED uses `Deep Supervised Network' (DSN) \cite{DSN} training to fine-tune the VGG network for the task of boundary detection, illustrated in \reffig{fig:hedtrain}. 
The principle behind  DSN can be loosely understood as classifier stacking  adapted to deep learning and turns out to be  practically very successful: if a multi-layer architecture is optimized for a given task, one can anticipate better results by informing each layer about the final objective, rather than relying on the final layer to back-propagate the information to its predecessors. This was shown to systematically improve convergence and test performance, both in generic detection tasks \cite{DSN} and in particular in the context of boundary detection \cite{HED}.

\newcommand{\ww}{\mathbf{W}}
\newcommand{\hh}{\mathbf{h}}
\newcommand{\wm}[1]{\mathbf{w}^{(#1)}}
\newcommand{\wmm}{\mathbf{w}}
\newcommand{\Loss}{\mathcal{L}}
\newcommand{\loss}{{l}}
\newcommand{\ff}{\mathbf{f}}

In particular, using the notation of \citet{HED}, we have a training set $S = (X_n,Y_n), n = 1,\ldots,N$  with $X_n$ being the input image and $Y_n = \{y^{(n)}_j, j=1,\ldots,|X_n|\},y^{(n)}_j \in \{0,1\}$ being the predicted labels (we will drop the $n$ subscript for brevity). 

We consider a multi-layer network, represented in terms of the union of its individual layer parameters, $\ww$, to which we append a set of per-layer `side' parameters $\wm{1},\ldots\wm{M}$. These side parameters aim at steering the intermediate layers of the network to extract features that are useful for the  classification task even when used on their own. 
 
The objective function of DSN/HED is phrased as:
\ba
\Loss_{side}(\ww,\wmm) = \sum_{m=1}^{M} \alpha_m \loss^m(\ww,\wm{m}),
\ea
where $\loss^m$  are the side-layer losses on the side output of the $m$-th layer and  $\alpha_m$ indicates the importance of the different side layer losses - e.g. setting $\alpha_m=0,m<M$, which amounts to standard training with a single loss at the top. 
In HED $\loss^m$ is a class-balanced cross-entropy loss: 
\begin{gather}
\!\!\!\!\!\loss^m(\ww,\wm{m}) \!=\! - \beta \sum_{j \in Y_{+}}\!\! \log P(y_j=1|X;\ww,\wm{m})\! - \!(1-\beta)\! \sum_{j \in Y_{-}}\!\! \log\!P(y_j=0|X;\ww,\wm{m})\!\! \label{eq:dsn}\\
 \doteq \sum_{j\in Y} w_{\hat{y}_j} S(\hat{y}_j,s_j^m), \label{eq:myeq}
\end{gather}
where \refeq{eq:dsn} $Y_{+},Y_{-}$ are the positive and negative training sample indices respectively, and $\beta$ is a design parameter set to mitigate the substantially larger number of negative samples in images. The probabilities in \refeq{eq:dsn} are obtained in terms of a sigmoidal function operating on the inner product $s_j^m  =\langle\wm{m},\ff_j\rangle$ between the side layer parameters $\wm{m}$ and the  features $\ff_j$ of the DCNN at position $j$, $P(y_j=1|X;\ww,\wm{m}) = \sigma(s_j^m)$.
In \refeq{eq:myeq} we  rewrite \refeq{eq:dsn} in a more general form where we sum over the whole image domain and use the ground truth label ${\hat{y}_j}$ to indicate which weight
and which of the two loss terms
is used per pixel $j$. 

An additional processing step of HED is a late fusion stage where the side outputs are combined into a final classification score. This is very meaningful for the task of boundary detection, as it exploits the multi-scale information extracted by the different processing layers of the DCNN. In particular, denoting by  ${S}^m$ the field of values predicted by the $m$-th side-layer, these are linearly combined into a final score, $S^{fs}  = \sum_{m=1}^M h_m S^{m}$; a fusion loss is used to learn the weights $\hh$ by calibrating the relative importance of the different side-layers when forming the final prediction:
\ba
\Loss_{fuse}(\ww,\wmm,\hh)   = \sum_{j\in Y} w_{\hat{y}_j} S(\hat{y}_j,\sum_{m=1}^M h_m s_j^m)
\ea

The overall objective function of HED is written as follows:
\ba
\Loss_{HED}(\ww,\wmm,\hh) = \Loss_{side} (\ww,\wmm) + \Loss_{fuse}(\ww,\wmm,\hh) \label{eq:hedtotal}
\ea
and is optimized using common Stochastic Gradient Descent training with momentum.

\section{Improved deep  boundary detection training}
\newcommand{\id}{55}
 \begin{figure}
 \vspace{-.2cm}
\begin{tabular}{c@{}c@{}c@{}c@{}c@{}c@{}c@{}c@{}c@{}c}
\begin{tabular}{c}
 \includegraphics[width=\scln\linewidth]{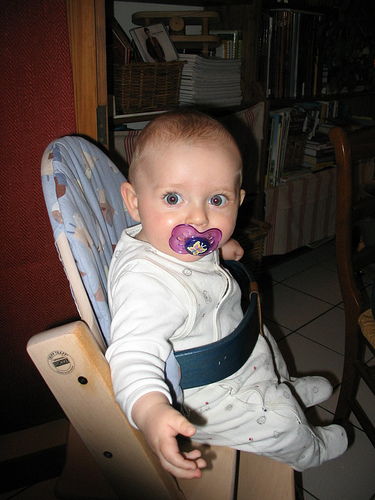}\\
  \includegraphics[width=\sclhl\linewidth]{\imroot/\id_im.png}\\
  \includegraphics[width=\sclq\linewidth]{\imroot/\id_im.png}
  \end{tabular}&
  \begin{tabular}{c}
  \includegraphics[width=\scln\linewidth]{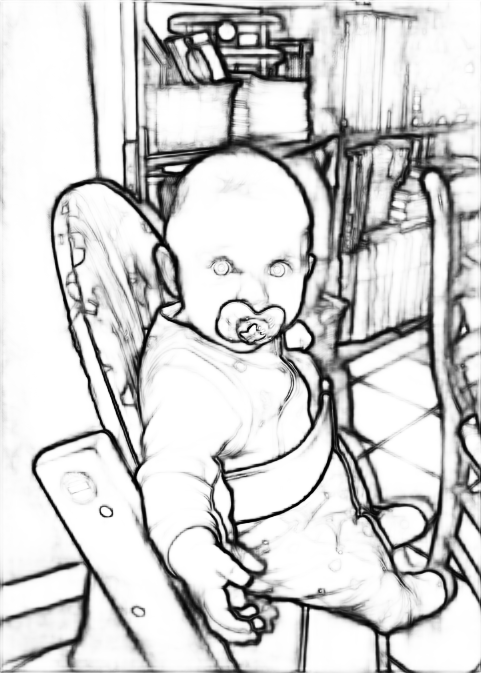} \\
 \includegraphics[width=\sclhl\linewidth]{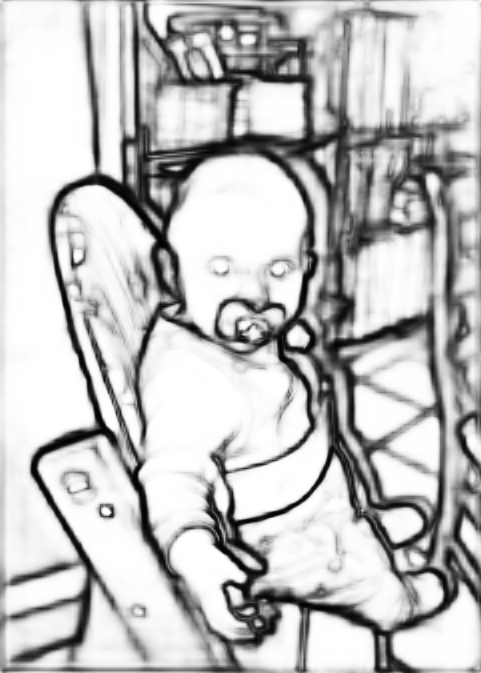} \\
 \includegraphics[width=\sclq\linewidth]{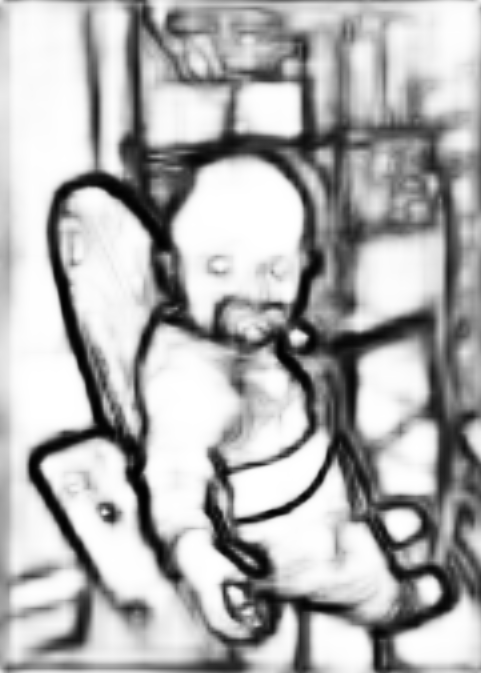}
 \end{tabular}& 
   \begin{tabular}{c}
   \\\\
  \includegraphics[width=\scln\linewidth]{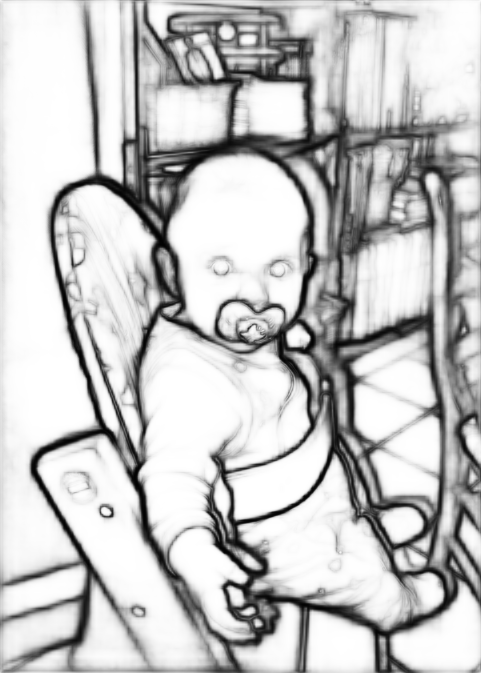} 
\end{tabular}
   & 
   \begin{tabular}{c}
 \includegraphics[width=\scln\linewidth]{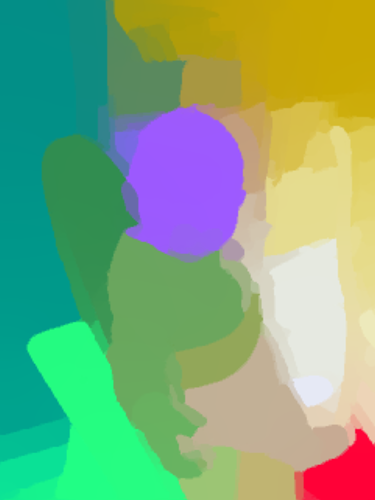} \\ 
  \includegraphics[width=\scln\linewidth]{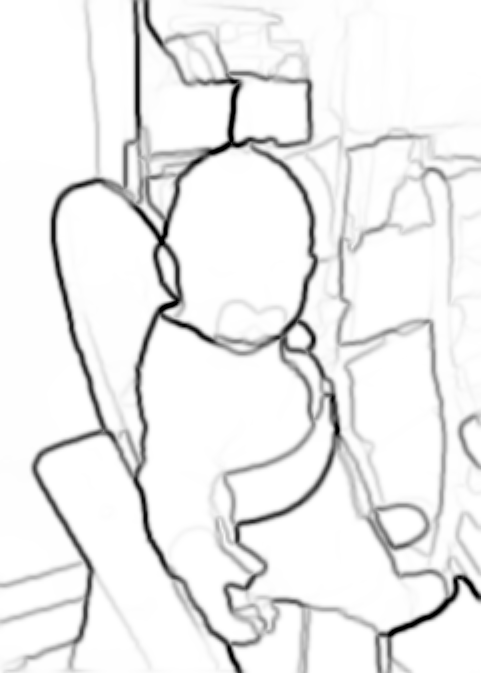} 
  \end{tabular}& 
       \begin{tabular}{c}
   \includegraphics[width=\scln\linewidth]{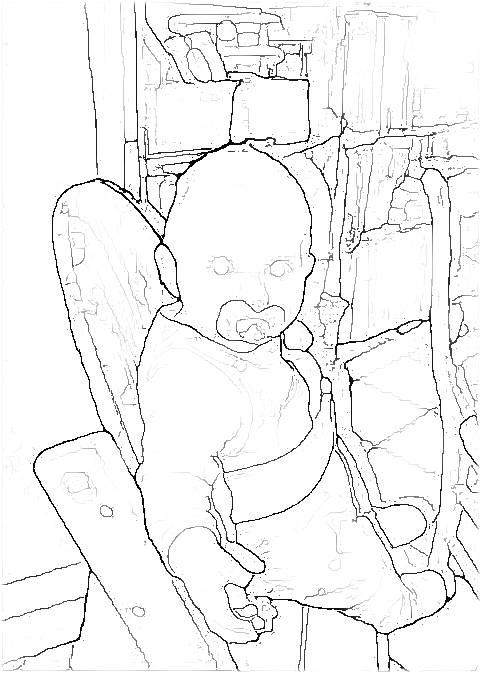}        \\
  \includegraphics[width=\scln\linewidth]{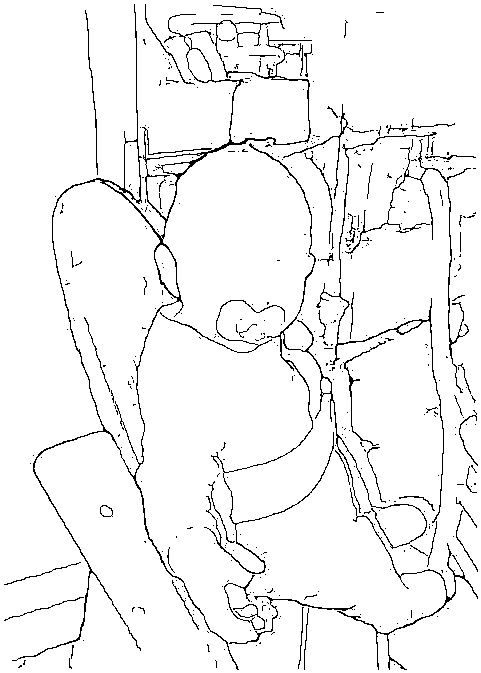} 
    \end{tabular}  
  \\
Image Pyramid & Tied CNN outputs & Scale fusion & NCuts \&  boundaries  & Final outputs
 \end{tabular}
\vspace{-.1cm}
 \caption{Overview of the main computation stages in our system: an input image is processed at three different scales in order to obtain multi-scale information. The the three scales are fused and sent as input to the  Normalized Cuts algorithm, that delivers eigenvectors (we show the first three of eight dimensions as an RGB image) and the resulting `Spectral Boundaries'. The latter are fused with the original boundary map, nonmaximum suppressed, and optionally thresholded (bottom row). All stages are implemented in Caffe, requiring less than a second on an Nvidia Titan GPU. \label{fig:system}\vspace{-.3cm}}
 \end{figure}
 
Having outlined the HED framework, we now turn to our contributions, consisting in (i) Multiple Instance Learning for boundary detection (ii) Graduated Deep Supervision (iii) Multi-Scale training, as well as introducing external data. 

 The improvements due to these contributions are summarized in \reftable{results}, where we report our ODS- and OIS-based F-measures on the BSD test set, alongside with the average precision (AP). We compare to our own HED-type baseline that yields a performance marginally below that of the original HED system of \citet{HED}; the latest system of \citet{HED} has an improved F-measure of $F=0.79$, due to additional dataset augmentation, which we have not performed yet.  We anticipate that this could further boost our already substantially better performance of $F=0.813$. Further comparisons can be found in Table.~2.
 
\subsection{Dealing with annotation inconsistencies}
\label{loss}

\begin{figure}[!b]
 \vspace{-.1cm}
 \centerline{	\includegraphics[width=.2\linewidth]{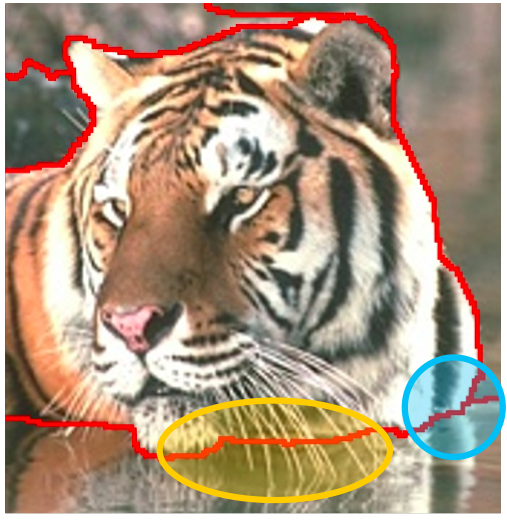}
 		\includegraphics[width=.2\linewidth]{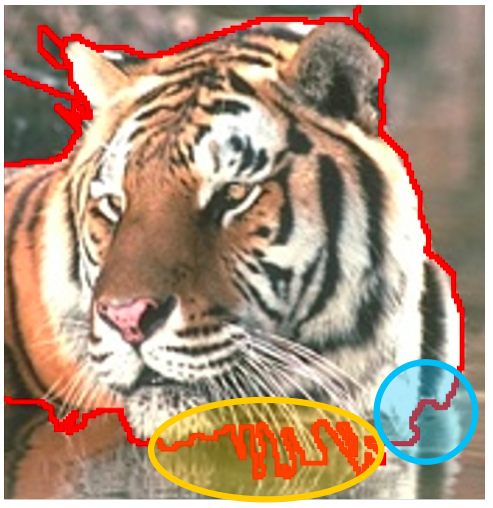}
}
\vspace{-.1cm}
 \caption{Location uncertainty of human annotations in the BSD dataset: even if annotators agree on the semantics, the  boundaries positions remain uncertain. As shown by the blue  circle, the precise position is unclear, while as shown by the orange ellipse, even the overall  boundary shape  may vary.
 	 \label{fig:lion}} \vspace{-.2cm}
 \end{figure}

  The first of our contributions  aims at dealing with  the inconsistency of human annotations in the BSD, illustrated in \reffig{fig:lion}. 
As can be seen, even if the two annotators agree about the semantics (a tiger in water), they may not place the boundaries at a common location. This makes it challenging to define `positive' and `negative' training samples in the vincinity of boundaries. 

This problem has already been acknowledged in the literature; for instance \citet{FuaStudent} turn boundary detection into a regression problem, by explicitly manipulating the ground truth to become smoother - which however may come at the cost of localization accuracy. In \citet{HED} a heuristic that was used was to only consider a pixel as positive if it is annotated consistently by more than three annotators. It is however unclear why other pixels should be labelled as negatives. 

Our approach builds on \citet{Kokkinos10}, where Multiple Instance Learning (MIL) \cite{diet97} is used to accommodate orientation inconsistencies during the learning of an orientation-sensitive boundary detector. That work was aimed at learning orientation-sensitive classifiers in the presence of orientation ambiguity in the annotations - we take a similar approach in order to deal with positional ambiguity in the annotations while 
learning a position-sensitive detector.  

 
Standard, `single instance' learning  assumes training samples come in feature-label pairs -or, as in  HED above, every pixel is either a boundary or not. Instead, MIL takes as a training sample a set of features (`bag') and its label. A bag should be labelled positive if at least one of its features is classified as positive, and negative otherwise. 

 In particular, since  human annotations come with some positional uncertainty, the standard evaluation protocol of \citet{MFM04} allows for some slack in the predicted position of a pixel (a fixed fraction of the image diagonal). One therefore does not need to label every positive pixel as a positive, but rather give a large score to a pixel in its vicinity - and to be more precise, a set of pixels in the line perpendicular to its orientation. 
 This set of pixels forms the bag associated to every positive pixel annotation. A pixel is declared negative if it is not contained in any positive bag. 
 
 More specifically, we associate every ground-truth boundary position  $j$ with a set  of $N_j$ positions and an associated feature bag, 
$\mathcal{X}_j =\{X_{j,1},\ldots,X_{j,N_j}\}$.  
These positions are estimated by  identifying the image positions that (i) lie closer to $i$ than any other ground-truth pixel and (ii) have a distance below a threshold $d$.

For each feature $X_{j,k}$ of the $j$-th bag our classifier provides a probability $p_{j,k}$ of it being positive, exactly as described in\refeq{eq:dsn}  but now the decision is taken by maximizing over instance probabilities: 
\ba
p_{\mathcal{X}_j} = P(y_j=1|\mathcal{X}_j) = \max_{k\in [1,\ldots N_j]}  p_{j,k} \label{eq:noisyor}
\ea
\newcommand{\bag}{\mathcal{B}}
The cost function now writes: 
\begin{gather}
\loss^m(\ww,\wm{m}) = \sum_{j\in Y_{-}} w_{\hat{y}_j} S(-1,s_j^m) +\sum_{j\in Y_{+}} w_{\hat{y}_j} S(1,\max_{j \in \bag_i} s_j^m)  
\end{gather}
where $\bag_i$ is the `bag' of pixel indices associated with sample $i$;
 this allows positive samples to select the neighbours that most support them while forcing all negatives to be negative. 
In terms of optimization, the $\mathrm{max}$ operation in \refeq{eq:noisyor} is not differentiable, but  we can  use a subdifferential of $p_j$:
\ba
\partial p_j = \frac{d p_{j,k^{\ast}}}{d f(X_{j,k^{\ast}})}, \quad\mathrm{where}~ k^{\ast} = \arg\max_k p_{j,k}.
\ea

The `MIL' column of 
\reftable{results} reports improvements over the baseline obtained by setting the distance, $d$ to 1; setting $d=2$ yields similar improvements.

\begin{table}
\vspace{-.7cm}
\begin{center}
\begin{tabular}{|l|c|c|c|c|c|c|}
\hline
Method & Baseline & MIL &  G-DSN  &  M-Scale  &  VOC  & Grouping \\
\hline
ODS & 0.7781     & 0.7863 &	0.7892   &   0.8033  &  0.8086 &  0.8134 \\
\hline
OIS &  0.7961   & 0.8083 &  0.8106 &   0.8196 &   0.8268 &   0.8308 \\
\hline
AP &   0.804    & 0.802 &  0.789  &   0.8483  &  0.861 &  0.866 \\
\hline
\end{tabular}
\end{center}
\caption{Improvements  obtained in this work over our own reproduction of a HED-type baseline : each column corresponds to a Section (MIL: \ref{loss}, G-DSN: \ref{gdsn}, Multi-Scale: \ref{mscale}, 
VOC: \ref{voc}, Grouping: \ref{grouping}).
Each improvement builds on top of the directly previous one. As performance indexes we report the `Optimal Dataset Scale' (ODS) F-measure (using a fixed detector threshold for whole dataset), the `Optimal Image Scale' (OIS) F-measure (using an oracle-based, image-dependent threshold), and `Average Precision' (AP). 
  \label{results}}
\end{table}

\subsection{Graduated DSN Training}
\label{gdsn}

The two terms in the objective function of HED, \refeq{eq:hedtotal}:
\ba
\Loss(\ww,\wmm,\hh) = \Loss_{side}(\ww,\wmm) + \Loss_{fuse}(\ww,\wmm,\hh)
\ea
 play a complementary role: the first, side-layer, terms force the intermediate layers to be discriminative and also extract some preliminary classification information; the second, fusion-layer, term calibrates the importance of the intermediate classifications delivered by the side-layers. 
 
As  discussed in \citet{DSN}, DSN can be understood as simplifying the associated learning problem in terms of optimization. But once the  network parameters are in the right regime, we can discard any simplifications that were required to get us there. This is a strategy used in the classical Graduated Non-Convexity  technique \cite{gnc}, and here we show  that it also helps improve DSN when applied to boundary detection.

\mycomment{
 our  observation is  that once the intermediate features have become discriminative only the second term is really needed - as the final decision is based on the fused classifier score, rather than the intermediate ones. Once we have reached  that stage it may actually be better to discard the side-layer losses while retaining the side-layer processing; 
}

For this we  modify  the training objective by associating the `side' term with a temporally decreasing weight while keeping the second term's weight fixed:
\ba
\Loss^{(t)}(\ww,\wmm,\hh) = (1 - \frac{t}{T}) \Loss_{side}(\ww,\wmm) + \Loss_{fuse}(\ww,\wmm,\hh), \nonumber
\ea
where $t$ is the current training epoch and $T$ is the total number of epochs.
Our training criterion  starts from DSN, where every intermediate layer is trained for classification, and eventually leads  to a skip-layer architecture, where the early layers are  handled exclusively by the final fusion criterion. By the end the fusion-layer can use the side-layers at will, without the compromises needed to keep the side losses low.  The improvements are reported in the G-DSN column of \reftable{results}.




\subsection{Multi-Resolution Architecture}
\label{mscale}

The authors of HED use `Deep Supervised Network' (DSN) \cite{DSN} training to fine-tune the VGG network for the task of boundary detection, illustrated in \reffig{fig:hedtrain}. However, image boundaries reside in multiple image resolutions  \cite{Witk83} and it has repeatedly been shown that fusing information from multiple resolutions improves boundary detection, e.g. in  \citet{sfs,berkeley11}. Even though the authors of  HED 
use information from multiple scales by fusing the outputs of many layers,  multi-resolution detection can still  help.

We first observed that simply averaging the results of the network applied to differently scaled versions of the image improved performance substantially, but then turned to a more accurate  way of doing the multi-resolution detection. As illustrated in \reffig{fig:hedtrain}, we consider a DSN-type multi-resolution architecture with tied weights, meaning that layers that operate at different resolutions share weights with each other. Parameter sharing across layers both accelerates convergence and also avoids over-fitting. We initialize the weights from a single-resolution architecture and fine-tune with a smaller set of iterations.
In order to capture fine-level boundaries the top-resolution image is an upsampled version of the original - e.g. for a $381 \times 421$ image from the BSD dataset we use a $577 \times 865$ upsampled version, from which we compute a three-level pyramid by downsampling by a factor of 2 and 4. 
The multi-resolution results are fused through an additional fusion layer that combines the fused results of the individual resolutions. The improvements are reported in the $S=3$ column of \reftable{results}.

 \begin{figure*}
 \vspace{-1cm}
 \centerline{	\includegraphics[width=.65\linewidth]{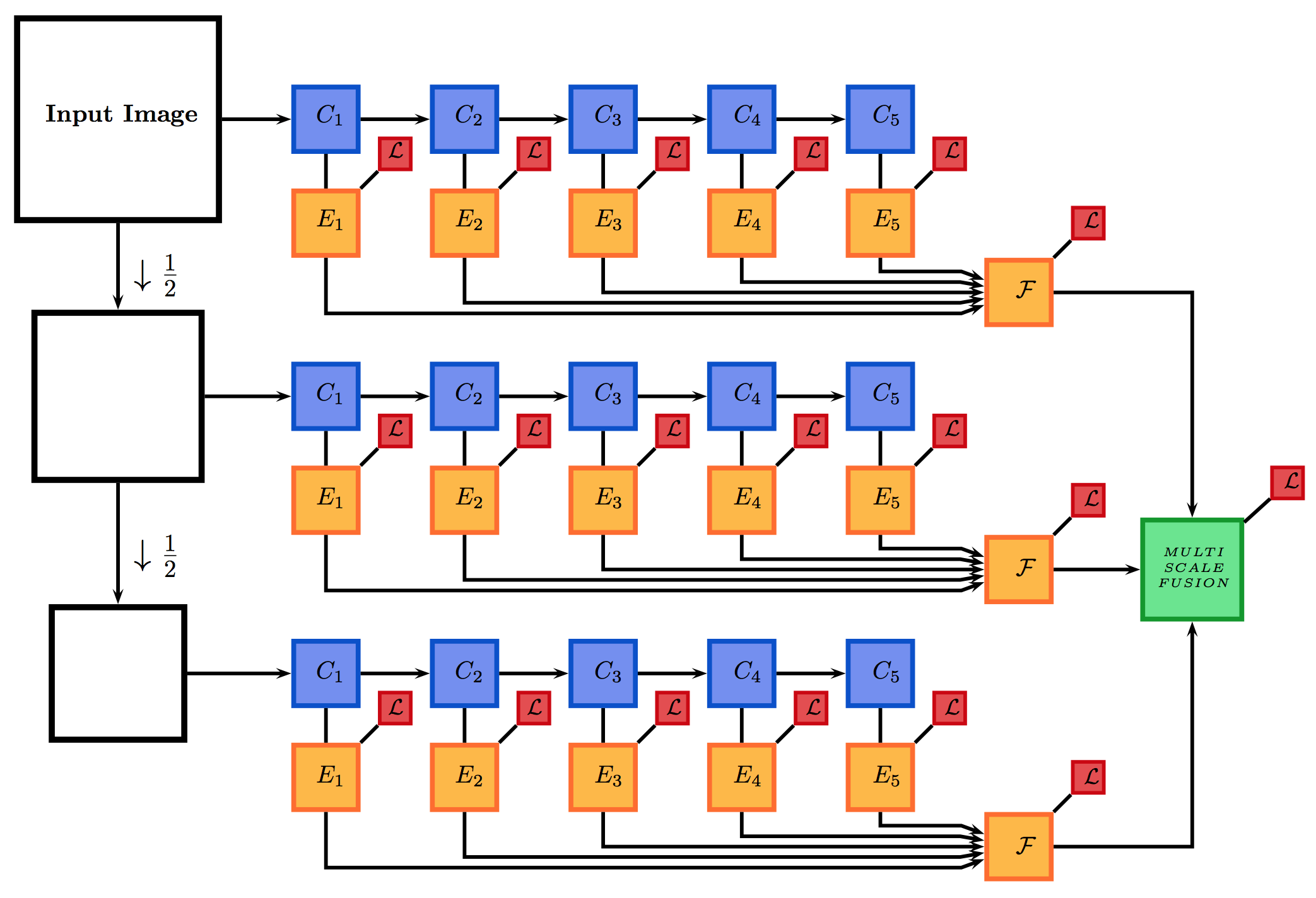}
}
 \vspace{-.1cm}
 \caption{Network architecture used for multi-resolution HED training: three differently scaled versions of the input image are provided as inputs to three FCNN networks that share weights - their multi-resolution outputs are fused in a late fusion stage, extending DSN to multi-resolution training.
 \vspace{-.5cm}
 	 \label{fig:mrtrain}}
 \end{figure*}

\subsection{Training with external data}
\label{voc}

Even though HED uses the pre-trained VGG network as initialization, dataset augmentation was reported to give substantial improvements. The authors in \citet{HED} originally used 32 geometric transformations (16 rotations and   flipping) of the 300 images used in the BSD trainval set, resulting in a total of roughly 10000 training images - in a recent version the authors consider two additional transformations are considered, resulting in roughly 30000 training images and pushing  performance from $F=0.78$ to $F=0.79$. 

We have not used these additional scalings in our experiments due to time constraints, but have considered  the use of boundaries from the VOC Context dataset \cite{context}, where all objects and `stuff' present in the scene are manually segmented. Our sole modification to those boundaries has been to label the interiors of houses as `don't care' regions that are ignored by the loss, since all of the windows, doors, or balconies that are missed by the annotators seemed to us as being legitimate boundaries.  
We only apply flipping to these images, resulting in roughly 20000 images, which are appended to the 10000 images we had originally used. 
As can be seen from the `VOC' column of \reftable{results}, this yields a substantial improvement.

\section{Using grouping in a deep architecture}
\label{grouping}

The combination of the techniques outlined above already help boundary detection outperform humans on the task of boundary detection - but still do not use any grouping information when delivering the probability of having boundaries. The boundary detector only implicitly exploits grouping cues such as closedness or continuity that can often yield improvements in the high-precision regime \cite{ZhuSS07,Kokkinos10b}. 

To capture such information we use the Normalized Cuts (NCuts) technique of \citet{Shi97,berkeley11}. We  treat the image as a weighted graph, where nodes corresponding to pixels and weights correspond 
to  low-level affinity between pixels measured in terms of the Intervening Contour cue \cite{Shi97}, where the contours are now estimated by our boundary detector. The NCut technique considers a relaxation of the discrete normalized cut optimization problem, which results in a generalized eigenvector problem \cite{Shi97}:
\begin{equation}
(D -W ) \mathbf{v} = \lambda D \mathbf{v},
\label{eq:overview:normalized_cuts}
\end{equation}
where $D$ is the graph degree matrix and $W$ is the affinity. The solutions to this generalized eigenvector problem can be understood \cite{laplacian} as euclidean embeddings of the inter-node distances - so nodes that have similar embeddings are likely to belong together and vice versa. 


One of the main impediments to the application of this technique has been computation time, requiring roughly 60 seconds on the CPU for a $321\times 481$ image for 10 eigenvectors. 
Even though accelerations exist, e.g. \citet{cour}, we found it simpler to harness the computational power of GPUs and integrate the Damascene system of \cite{damascene} with the Caffe  deep learning framework. The implementation of \cite{damascene} provides a GPU implementation of the Lanczos solver for the  generalized eigenvector problem of \refeq{eq:overview:normalized_cuts} that is two orders of magnitude faster than the CPU-based algorithm.  
When integrated with our boundary detector Damascene yields 8 eigenvectors for a  $577 \times 865$  image in less that 0.2 seconds. It is also straightforward to use a downsampled version of the boundary map to yield further accelerations. 

These embeddings can be used for boundary detection in terms of their directional derivatives, in order to provide some `global' evidence for the presence of a boundary, known as the `spectral probability of boundary' cue \cite{berkeley11}. In particular, as in \cite{berkeley11}, we obtain a new boundary map in terms of a linear combination between the posterior probabilities delivered by our multi-resolution network and the spectral boundary magnitude. This further improves the performance of our detector, yielding an F-measure of 0.813, which is substantially better than our earlier performance of 0.807, and  humans, who operate at 0.803. We anticipate that adding a few processing layers can further improve performance.

\begin{figure}[!htb]
\ffigbox{%
\includegraphics[width=.35\linewidth]{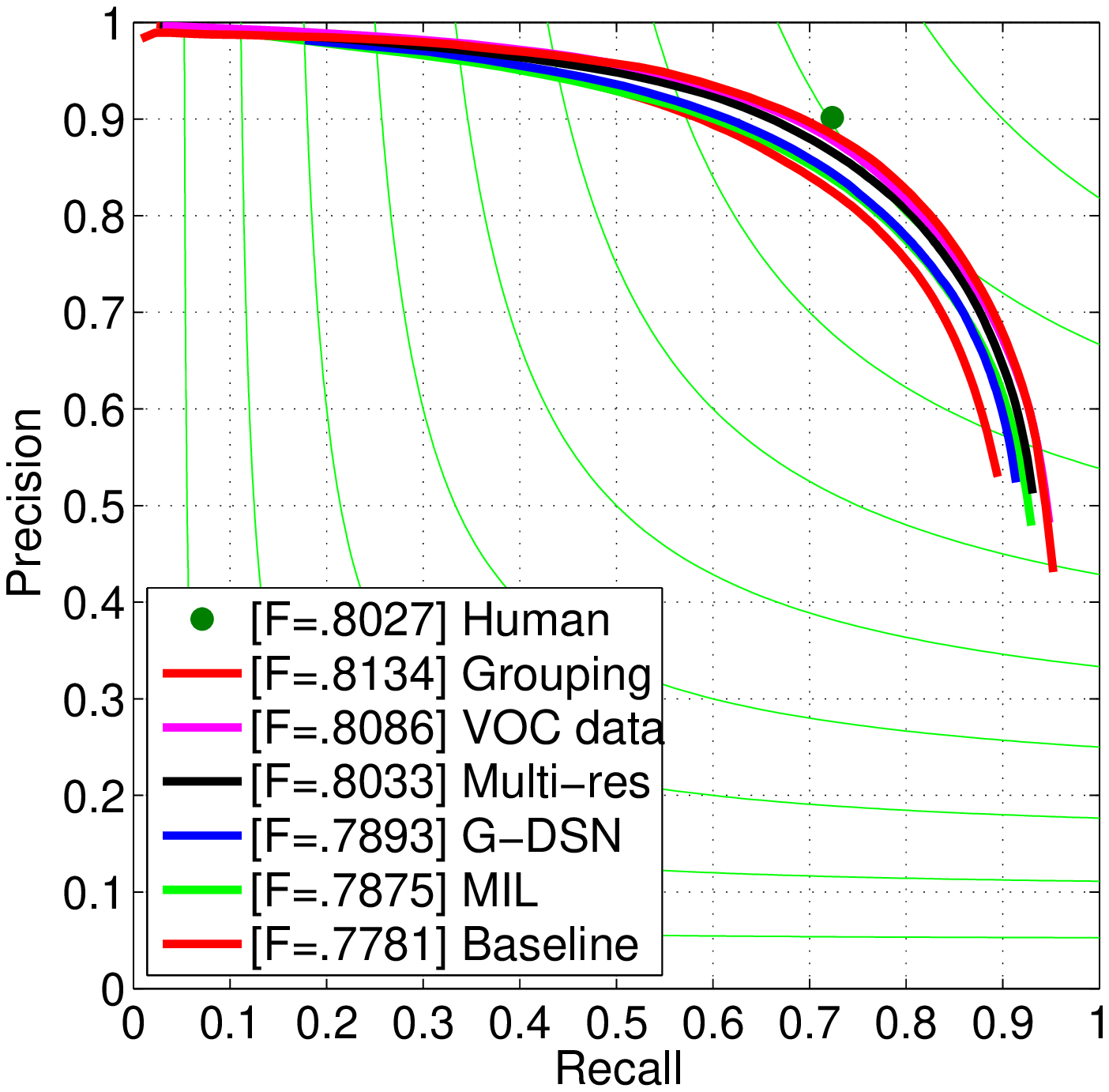}
\includegraphics[width=.35\linewidth]{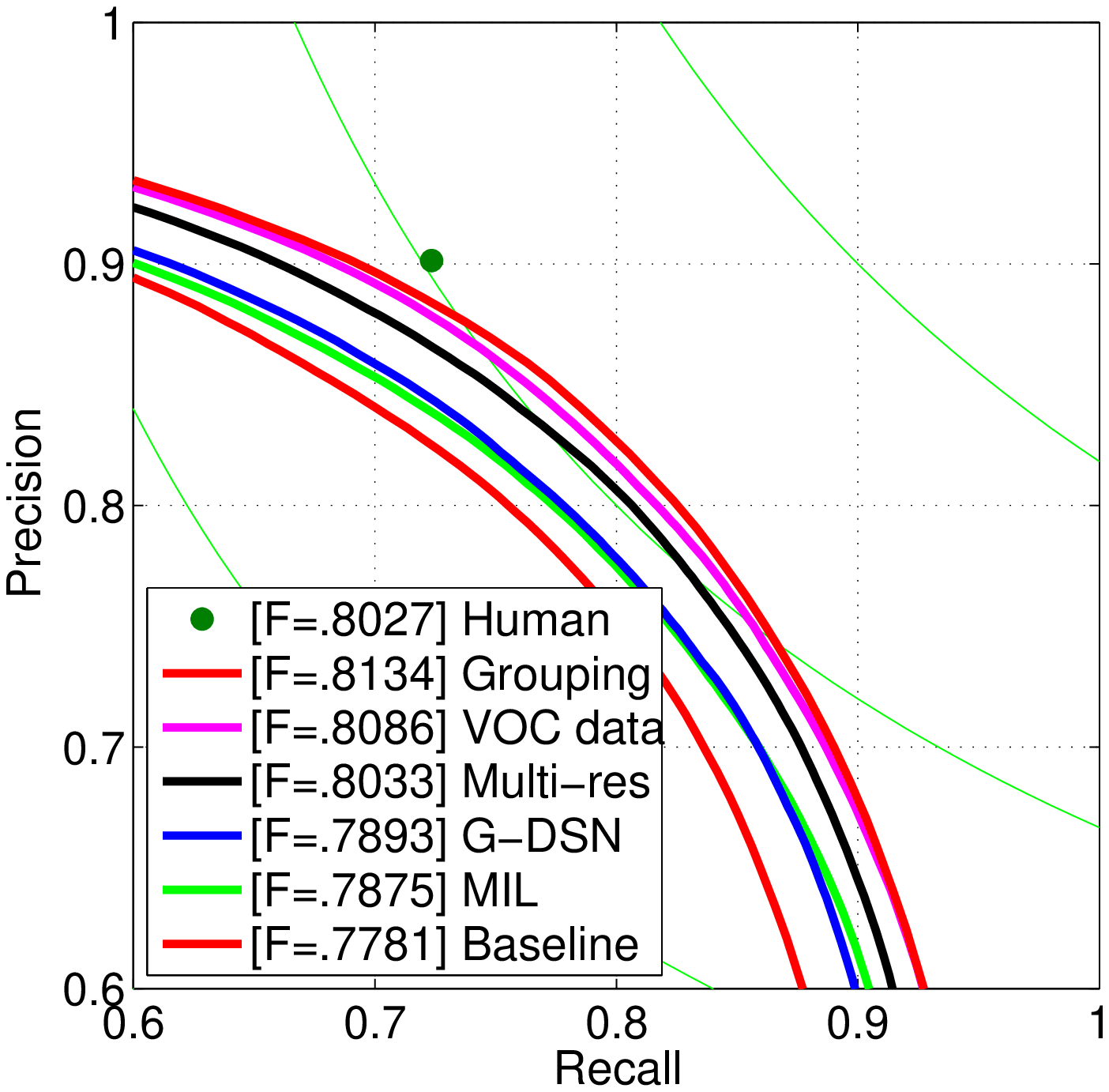}
}{%
  \caption{ Impact of the different improvements described in Section 2: starting from a baseline that performs only slightly worse than the HED system of \cite{HED} we end up with a detector that largely surpasses  human F-measure, illustrated in terms of green isocontours. On the right we zoom into the high-F measure regime.  \label{fig:plot}}%
}
\end{figure}

We summarize the impact of the different steps described above in \reffig{fig:plot} - starting from a baseline (that performs slightly worse than the HED system of \citet{HED} we have introduced a series of changes that resulted in a system that performs boundary detection with an F-measure that exceeds that of humans. When compared to the current state-of-the-art method of \citet{HED} our method clearly dominates in terms of all typical performance measures, as shown in Table \ref{table:hedfinal}.

Indicative qualitative results are included in the supplemental material.

\begin{table}
\resizebox{.6\linewidth}{!}{
\begin{tabular}{|@{}l@{}|c|c|c|}
\hline
Method  & ODS & OIS  & AP \\
\hline
\hline
gPb-owt-ucm \cite{berkeley11}   &  0.726 & 0.757 & 0.696 \\
SE-Var        \cite{sfs} &  0.746 & 0.767 & 0.803 \\
DeepNets      \cite{aistat} &  0.738 & 0.759 & 0.758 \\
N4-Fields     \cite{ganin2014n} &  0.753 & 0.769 & 0.784 \\
DeepEdge      \cite{shi15} &  0.753 & 0.772 & 0.807 \\
CSCNN         \cite{iclrbnd} &  0.756 & 0.775 & 0.798 \\
DeepContour  \cite{shen2015deepcontour}  &  0.756 & 0.773 & 0.797 \\
HED-fusion  \cite{HED} &  0.790 & 0.808 & 0.811 \\
HED-late merging  \cite{HED} &  0.788 & 0.808 & 0.840 \\
\hline
Ours (DCNN + sPb) & 0.8134 & 0.8308 & 0.866\\
\hline
\end{tabular}
}
\caption{Comparison to the state-of-the-art in boundary detection, including the latest version of HED, trained with its most recent dataset augmentation \cite{HED}.  We clearly outperform HED across all performance measures, while keeping the  speed above 1 frame per second. \label{table:hedfinal}}
\end{table}



\section{Synergy with semantic segmentation}
Having pushed the performance of boundary detection to a good level, we now turn to seeing how it can be explored in the context of the higher level task of semantic segmentation. 
\footnote{In an earlier version of this Arxiv report we had considered the combination with object proposals using the system of \citet{lpo} and reported  large improvements. This was due to an erroneous calculation of the baseline. After fixing the error there are still some improvements, but they are not large enough to be considered substantial.}

Since our model is fully-convolutional we can easily  combine it with the recent line of works around FCNN-based semantic segmentation\cite{LongSD14,Chen2015iclr,papa15,torr15}. These have delivered excellent results, and in particular the use of the Dense
Conditional Random Field (DenseCRF) of \citet{krahenbuhl2011efficient} by \citet{Chen2015iclr,papa15,torr15}, has enhanced the discriminative power of FCNNs with local evidence gathered by the image intensity. 

Following \citet{Chen2015iclr} we define the CRF distribution as:
\begin{align}
  P(\boldsymbol{x}) = \frac{1}{Z} \exp \Big(-E(\boldsymbol{x})\Big), \quad  E(\boldsymbol{x}) = \sum_i \phi_i(x_i)   + \sum_{ij} \theta_{ij}(x_i, x_j). \label{eq:crf}
\end{align}
where $\boldsymbol{x}$ is the pixel-label assignment and $E(\boldsymbol{x})$ is the energy function.
In \refeq{eq:crf} $ \phi_i(x_i) = - \log P(x_i)$ with $P(x_i)$ being the CNN-based probability of assigning label $j$ to pixel $i$, and $\theta_{ij}(x_i, x_j)$ is a bilateral filter-like  image-based pairwise potential between  $i$ and $j$:
\begin{gather}
  \theta_{ij}(x_i, x_j) = w^1 \exp \Big(-\frac{|p_i-p_j|^2}{2\sigma_\alpha^2} -\frac{|I_i-I_j|^2}{2\sigma_\beta^2} \Big)   + w^2 \exp \Big(-\frac{|p_i-p_j|^2}{2\sigma_\gamma^2}\Big). \label{eq:rgb}
\end{gather}
The first kernel in  \refeq{eq:rgb} depends on both pixel positions (denoted as $p$) and pixel color intensities (denoted as $I$), while the second kernel only depends on pixel positions - the hyper-parameters $\sigma_\alpha$, $\sigma_\beta$ and $\sigma_\gamma$ control the Gaussian kernels. Mean-field Inference for this form of pairwise terms can be efficiently implemented with high-dimensional filtering  \cite{adams2010fast}.

 
Our modifications are very simple:  firstly, we adapt the multi-resolution architecture outlined in the previous section to semantic segmentation. Using multi-resolution processing with tied-weights and performing  late score fusion yielded substantially better results than using a single-resolution network: 
as shown in \reftable{segtable} 
when combining the multi-scale network's output with  DenseCRF inference, performance increases from 72.7 (single-scale counterpart of \citet{Chen2015iclr}) or 73.9 (skip-layer multi-scale counterpart of \citet{Chen2015iclr}) to 74.8 (our multi-scale) in mean accuracy. 

Secondly, we integrate the boundary information extracted by our detector into the DenseCRF  by using the eigenvectors computed by normalized Cuts to augment the RGB color features of \refeq{eq:rgb}, thereby conveying boundary-based proximity into DenseCRF inference. In particular we augment the dimensionality of $I_i$ in \refeq{eq:rgb} from 3 to 6, by concatenating the 3 eigenvectors delivered by NCuts with the RGB values.
We observe that introducing the Normalized Cut eigenvectors into DenseCRF inference yields a clear improvement over an already high-performing system (from 74.8 to 75.4), while  a small additional improvement was obtained we performing graph-cut inference with pairwise terms that depend on the boundary strength (from 75.4 to 75.7). Further improvements can be anticipated though an end-to-end training using the recursive CNN framework of \citet{torr15}  as in the currently leading works - we will explore this in future work. 

Indicative qualitative results are included in the supplemental material.

\begin{table}[!htb]
\begin{center}
\resizebox{.6\linewidth}{!}{
\begin{tabular}{|l|c|}
\hline
Method  &  mAP $\%$ \\
\hline
Adelaide-Context-CNN-CRF-COCO \cite{lin2015efficient} & 77.8 \\
\hline
CUHK-DPN-COCO \cite{liu2015semantic} & 77.5  \\
\hline
Adelaide-Context-CNN-CRF-COCO \cite{lin2015efficient} & 77.2 \\
\hline
MSRA-BoxSup  \cite{dai2015boxsup} & 75.2 \\
\hline
Oxford-TVG-CRF-RNN-COCO  \cite{torr15} & 74.7 \\
\hline
DeepLab-MSc-CRF-LF-COCO-CJ  \cite{Chen2015iclr}& 73.9  \\
\hline
DeepLab-CRF-COCO-LF\cite{Chen2015iclr} & 72.7  \\ \hline
Multi-Scale DeepLab  & 72.1 	\\
Multi-Scale DeepLab-CRF  & 74.8 	\\
Multi-Scale DeepLab-CRF-Embeddings & 75.4 	\\
Multi-Scale DeepLab-CRF-Embeddings-GraphCuts & 75.7 	\\
\hline
\end{tabular}
}
\vspace{-.1cm}
\end{center}
\caption{Mean Average Precision performance on the VOC 2012 Semantic Segmentation test set; our results are in the last four rows. We start from a novel multi-resolution variant of  DeepLab and consider the gain of introducing normalized cut eigenvectors  into DenseCRF inference, as well as adding a more classical boundary-sensitive GraphCut post-processing stage on top.  \label{segtable}} 
\vspace{-.1cm}
\end{table}

\section{Conclusion}
We have proposed a method to substantially improve deep learning-based boundary detection performance. 
Our system is fully integrated in the Caffe framework and operates in less than one second per frame. Its F-measure, as measured on the standard BSD dataset is higher than that of humans.

We anticipate that further improvements can be gained through a joint treatment of other low-level cues,  such as  symmetry  \cite{TsogkasK12} or surface orientation, and depth \cite{eigen2014predicting}. We also intend to further explore the merit of our detector in the context of high-level tasks, such as  object detection and recognition. 

\section{Acknowledgements}
This work was supported by FP7-RECONFIG and equipment donated by NVIDIA. 
I thank the authors of \citet{HED} for inspiration, Alp Guler for  illustrations and tables, Kostas Papazafeiropoulos for help with porting Damascene to Caffe, George Papandreou for guidance on  Caffe and Pierre-Andr\'e Savalle for teaching me to handle prototxt files like a professional \verb+sed+user.


\newpage
\section{Supplemental Material}
We provide below qualitative results on images from the Pascal VOC test set. 
 
\renewcommand{\scln}{.18}

\newcommand{\putit}[1]{
\begin{tabular}{ccccc}
\includegraphics[width=\scl\linewidth]{\imroot/#1_im.png}
& 
\includegraphics[width=\scln\linewidth]{\imroot/#1_PE.png}
&
\includegraphics[width=\scln\linewidth]{\imroot/#1_eig.png}
& 
\includegraphics[width=\scln\linewidth]{\imroot/#1_seg_un.png}
&
\includegraphics[width=\scln\linewidth]{\imroot/#1_seg_emb.png}
\end{tabular}}

\newcommand{\putitf}[1]{
\begin{tabular}{ccccc}
\includegraphics[width=\scl\linewidth]{\imroot/#1_im.png}
& 
\includegraphics[width=\scln\linewidth]{\imroot/#1_PE.png}
&
\includegraphics[width=\scln\linewidth]{\imroot/#1_eig.png}
& 
\includegraphics[width=\scln\linewidth]{\imroot/#1_seg_un.png}
&
\includegraphics[width=\scln\linewidth]{\imroot/#1_seg_emb.png}
\\
(a)  & (b)  & (c) & (d) & (e)
\end{tabular}}

\begin{figure}[!b]
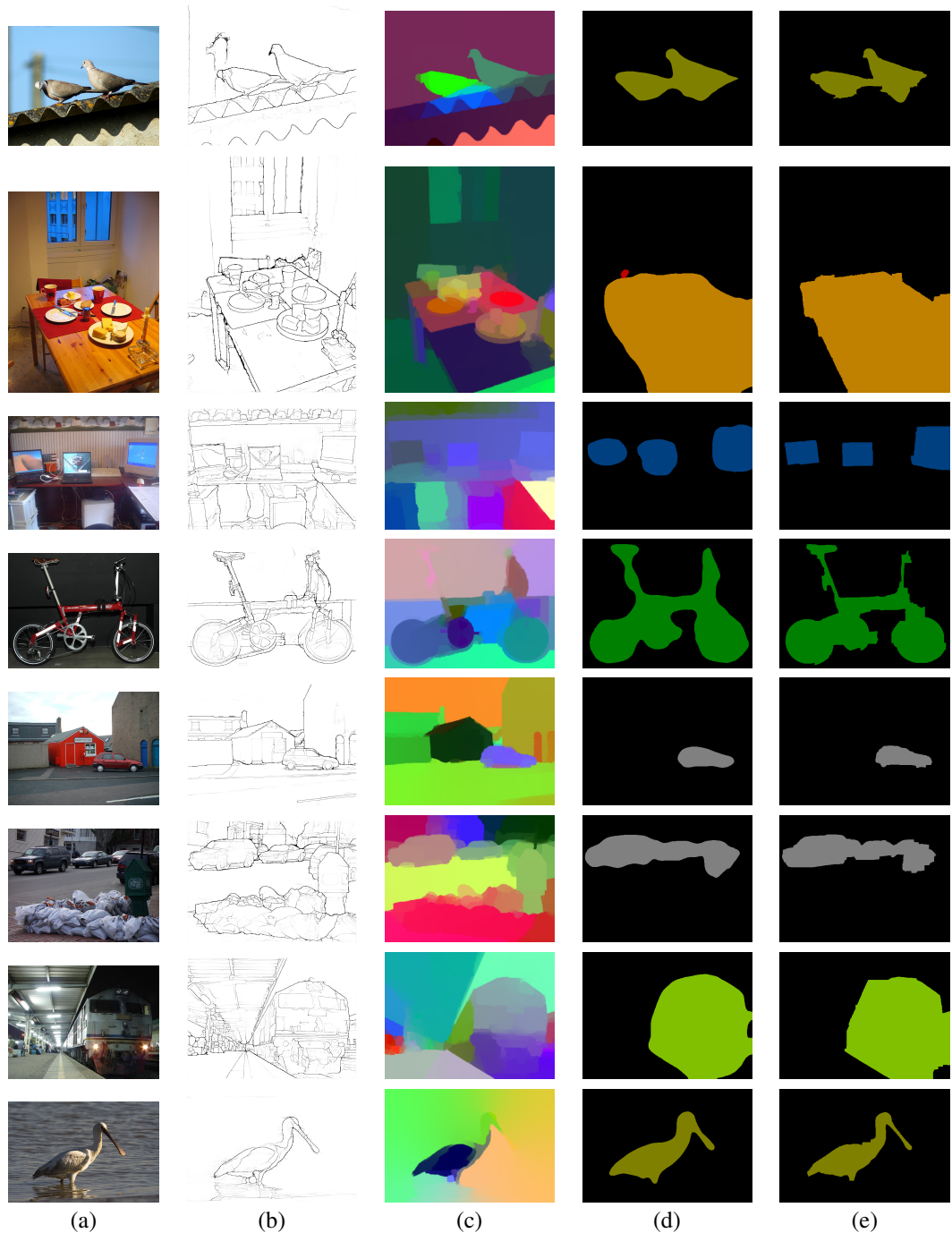

\putit{21}
\putit{36}
\putit{33}
\putit{38}
\putit{39}
\putit{2}
\putit{15}
\putitf{42}
\caption{Indicative results on the PASCAL VOC 2012 test set: for each image we show in (b) the final estimate of the probability of boundary, in (c) three  leading eigenvectors delivered by  Normalized Cuts (d) the semantic segmentation that would be obtained by our multi-scale DCNN variant of DeepLab, prior to DenseCRF inference and (e) the improved result obtained by combining DenseCRF inference with the normalized Cut embeddings and the image  boundaries. 
}
\end{figure}
\mycomment{
\begin{figure}[!b]
\putitf{43}
\putit{7}
\putit{12}
\putit{14}
\putit{16}
\putit{30}
\putitf{20}
\caption{Indicative results on the PASCAL VOC 2012 test set: for each image we show in (b) the final estimate of the probability of boundary, in (c) three  leading eigenvectors delivered from  Normalized Cuts (d) the semantic segmentation that would be obtained by our multi-scale DCNN variant of DeepLab, prior to DenseCRF inference and (e) the improved result obtained by combining DenseCRF inference with the normalized Cut embeddings and the image  boundaries. 
}
\end{figure}

}

\mycomment{
\newpage
\begin{figure}
\putit{32}
\putit{34}
\putit{35}
\putit{37}
\end{figure}
 \begin{figure}[!htb]
 \centerline{	
 \includegraphics[width=.12\linewidth]{6046.jpg}
 \includegraphics[width=.12\linewidth]{6046_bnd.jpg}
 \includegraphics[width=.12\linewidth]{196027.jpg}
 \includegraphics[width=.12\linewidth]{196027_bnd.jpg}
 \includegraphics[width=.12\linewidth]{164046.jpg}
 \includegraphics[width=.12\linewidth]{164046_bnd.jpg}
  \includegraphics[width=.12\linewidth]{104055.jpg}
}
 \centerline{	
 \includegraphics[width=.12\linewidth]{257098.jpg}
 \includegraphics[width=.12\linewidth]{257098_bnd.jpg}
 \includegraphics[width=.12\linewidth]{104055.jpg}
 \includegraphics[width=.12\linewidth]{104055_bnd.jpg}
 \includegraphics[width=.12\linewidth]{279005.jpg}
 \includegraphics[width=.12\linewidth]{279005_bnd.jpg}
  \includegraphics[width=.12\linewidth]{372019.jpg}
}
\caption{Indicative results from the BSD test set}
 \end{figure}}

\small
\newcommand{\noopsort}[1]{} \newcommand{\printfirst}[2]{#1}
  \newcommand{\singleletter}[1]{#1} \newcommand{\switchargs}[2]{#2#1}

\end{document}